\documentclass[final,5p,times,twocolumn, authoryear]{elsarticle}

\usepackage{amssymb}

\usepackage{hyperref}

\usepackage{float}
\usepackage{algorithm}
\usepackage{algorithmicx}
\usepackage{algpseudocode}
\algnewcommand\algorithmicparam{\textbf{Parameter:}}
\algnewcommand\Param{\item[\algorithmicparam]}
\usepackage{amsmath}
\usepackage{caption}
\usepackage{graphicx}
\usepackage{float} 
\usepackage{subcaption}
\usepackage{booktabs}
\usepackage{multirow}
\usepackage{svg}
\usepackage{threeparttable}
\usepackage{natbib} 
\usepackage{tabularx}
\usepackage[font=small,labelfont=bf,labelsep=none]{caption}  
\usepackage{ulem} 

\journal{Neural Networks}

\begin{document}

\begin{frontmatter}



\title{Brain-Inspired Efficient Pruning: Exploiting Criticality in Spiking Neural Networks}


\author[1,2,3]{Shuo Chen}
\author[4]{Boxiao Liu}
\author[1]{Zeshi Liu}
\author[1,3]{Haihang You\corref{cor1}}
\ead{youhaihang@ict.ac.cn}
\cortext[cor1]{Corresponding author}
\affiliation[1]{organization={State Key Lab of Processors, Institute of Computing Technology, Chinese Academy of Science},
Institute of Computing Technology
            city={Beijing},
            country={China}}
\affiliation[2]{organization={School of Computer Science and Technology, University of Chinese Academy of Sciences},
            city={Beijing},
            country={China}}
\affiliation[3]{organization={Zhongguancun Laboratory},
            city={Beijing},
            country={China}}
\affiliation[4]{organization={Sensetime Research},
            city={Beijing},
            country={China}}
\begin{abstract}

Spiking Neural Networks (SNNs) have gained significant attention due to the energy-efficient and multiplication-free characteristics. Despite these advantages, deploying large-scale SNNs on edge hardware is challenging due to limited resource availability. Network pruning offers a viable approach to compress the network scale and reduce hardware resource requirements for model deployment. However, existing SNN pruning methods cause high pruning costs and performance loss because they lack efficiency in processing the sparse spike representation of SNNs. In this paper, inspired by the critical brain hypothesis in neuroscience and the high biological plausibility of SNNs, we explore and leverage criticality to facilitate efficient pruning in deep SNNs. We firstly explain criticality in SNNs from the perspective of maximizing feature information entropy. Second, We propose a low-cost metric for assess neuron criticality in feature transmission and design a pruning-regeneration method that incorporates this criticality into the pruning process. Experimental results demonstrate that our method achieves higher performance than the current state-of-the-art (SOTA) method with up to 95.26\% reduction of pruning cost. The criticality-based regeneration process efficiently selects potential structures and facilitates consistent feature representation.

\end{abstract}




\end{frontmatter}

\section{Introduction}

Spiking Neural Networks (SNNs) have received significant attention in recent years~\citep{wu2019direct,zheng2021going} as the third generation of neural networks~\citep{maass1997networks}. SNNs are widely employed in resource-constrained hardware~\citep{akopyan2015truenorth, davies2018loihi, pei2019towards, sushi}, relying on energy-efficient and multiplication-free advantages from sparse spike signals. 

The limited computational and storage resources of hareware pose challenges to implementing deep SNNs with large-scale parameters.  Network pruning provides a potential solution to address resource issues. There have been a lot of efforts to unstructured pruning for SNNs. Methods inspired by the human brain include modeling synaptic regeneration processes~\citep{kundu2021spike} and dendritic motion~\citep{kappel2015network}. Recent efforts reference to techniques from Artificial Neural Network (ANN) pruning. \citet{kim2022exploring} explored the ``lottery ticket" hypothesis in SNN pruning with Iterative Magnitude Pruning (IMP). \citet{deng2021comprehensive} combined Spatio-Temporal Backpropagation (STBP) and Alternating Direction Method of Multipliers (ADMM) and proposed an activity-based regularization method. Grad R~\citep{ijcai2021-236} improved the Deep R method by introducing a weight regeneration mechanism. RCMO-SNN~\citep{chen2023resource} proposed an end-to-end Minimax optimization method for sparse learning.

\begin{table}[htb]
\caption{ Test accuracy and required pruning epochs on ResNet19 with about 95\%  sparsity for CIFAR100 using different pruning methods.}
\begin{center}
\begin{tabular}{lccc}
\toprule
 \textbf{Method}  & \textbf{\#Epoch} & \textbf{Acc. }(\%)  \\
\midrule
Grad R~\citep{ijcai2021-236}  & 2048 & 67.47 \\
IMP~\citep{kim2022exploring}  & 3380 & 70.54 \\
RCMO-SNN~\citep{chen2023resource}  & 3100 & 72.67 \\
NDSNN~\citep{dac2023}  & 300 & 68.95 \\
This work  & 200 & 73.01  \\
\bottomrule
\end{tabular}
\label{tab:intra}
\end{center}
\end{table}

However, current state-of-the-art (SOTA) pruning methods spend significant time costs in the SNN pruning process compared to ANNs. Table \ref{tab:intra} summarizes the required epochs and pruned accuracy on ResNet19 for CIFAR100 using different pruning methods. Existing SOTA methods ~\citep{ijcai2021-236, kim2022exploring, chen2023resource} require over 2000 epochs to achieve approximately 95\% sparsity on ResNet19.  Efforts to reduce pruning costs include NDSNNs based on sparse training \citep{dac2023}, which reduce the epoch requirement to 300 but with a significant loss of accuracy. Other approaches ~\citep{kim2022exploring, kim2023exploring} attempt to reduce pruning costs by compressing time steps during the pruning process, but these methods only achieve limited acceleration ratios (1.12 to 1.59) and still result in significant performance losses. 

The high pruning cost comes from the multiple rounds of pruning and long fine-tuning time required by SOTA SNN pruning methods to achieve high-performance pruned models. Unlike the continuous value processing in ANNs, SNNs transmit information using sparse spike signals, relying on specialized neuron activations and neural signal encoding, which poses challenges for learning spike feature patterns in deep SNNs~\citep{zheng2021going, wu2019direct, shrestha2021hardware}. The multiple rounds of pruning and fine-tuning exacerbate the learning challenge. Specifically, it requires the long recovery time after each pruning round~\citep{kim2022exploring, chen2023resource}, or forces a reduction in the magnitude of single pruning step for rapid recovery~\citep{ijcai2021-236}. Both lead to a multiplication of pruning costs, accompanying a loss in accuracy. Therefore, it is necessary to develop a new SNN pruning method that can effectively preserve the critical spike patterns and reduce fine-tuning costs without sacrificing performance.

In this paper, inspired by the critical brain hypothesis in neuroscience, we explore and leverage criticality to facilitate efficient pruning in deep SNNs. The critical brain hypothesis~\citep{turing2009computing} suggests that neurons operating at a critical state possess excellent information transmission capabilities, crucial for complex feature extraction and learning in the brain~\citep{kinouchi2006optimal, beggs2008criticality, shew2009neuronal, beggs2012being}. SNNs mimic the mechanisms of biological neural systems and demonstrate high biological plausibility. So we employ the criticality in SNN pruning, aiming to maximally preserve the feature extraction capability in pruned models and accelerate the entire pruning process.

We first analyze criticality in SNNs through information entropy, finding that neurons at critical states can encode and transmit information more efficiently. By combining the principle of maximizing feature information entropy and observations on the behavior of neurons at critical states \citep{gal2013self}, we propose a criticality metric based on the derivative of neuron surrogate functions. This metric assesses the criticality of neurons in information processing while avoiding additional computational costs. To integrate this criticality into the pruning process of SNNs, we design a pruning-regeneration method. Experimental results show that our method, evaluated on both unstructured and structured pruning, demonstrates higher performance and lower pruning costs compared with state-of-the-art methods. The criticality metric effectively identifies and preserves critical structures. Our method achieves lower intra-cluster variance within the class and higher feature cosine similarity between the training samples and test samples, confirming the effectiveness of the criticality mechanism in enhancing feature extraction and consistency of pruned models.

In summary, our key contributions are as follows:
\begin{itemize}
\item Inspired by the critical brain hypothesis in neuroscience, we propose a metric for the neuron criticality of SNNs and design a pruning-regeneration method to improve feature extraction and achieve efficient pruning for SNNs.
\item We verify the utility of neuron criticality and its metric in SNNs from the perspective of maximizing feature information entropy. Furthermore, Experimental results show that the proposed criticality metric efficiently selects potential structures, reduces intra-cluster variance in 93\% of classes on average, and increases feature cosine similarity between training and test samples in over 98\% of classes. These findings highlight improved feature extraction and consistency in pruned models.
\item We evaluate our method for unstructured pruning. Our method achieves higher performance compared to SOTA methods on VGG16 and ResNet19 for CIFAR100 with from 91.15\% to 95.25\% running cost reduction for over 90\% sparsity. 
\item We perform the structured pruning for deep SNNs using our method, yielding better results than the sophisticated SOTA method. To our knowledge, this is the first work on structured pruning with high floating point operations (FLOPs) reduction ($>$50\%) in deep SNNs.
\end{itemize}

\section{Related Work}
\subsection{Network Pruning}
Pruning refers to removing specific structures from a network to induce sparsity. Unstructured pruning~\citep{han2015learning} removes weight parameters to achieve a high level of connection sparsity. However, standard hardware does not fully optimize sparse matrix operations, limiting potential acceleration from unstructured pruning. Structured pruning~\citep{he2017channel} achieves structured sparsity by removing entire kernels or channels. This approach is hardware-friendly but might not reach the highest sparsity.

\subsubsection{Pruning Criterion}
The most commonly employed pruning paradigm involves evaluating the impact of structure on network performance through a pruning criterion and removing the least significant ones. For unstructured pruning, the magnitude of each weight has emerged as the most widespread pruning criterion, which was firstly proposed in~\citep{han2015learning} and has seen widespread adoption in pruning for ANNs~\citep{zhu2017prune, liu2021sparse} and SNNs~\citep{kim2022exploring, dac2023}. The synaptic parameter~\citep{bellec2017deep}, which represents the connection strength, is another pruning criterion.~\citet{ijcai2021-236} improved the synaptic parameter with gradient rewiring and applied it to SNN pruning. For structured pruning, the first and second-order information of the gradient is utilized to design importance scores~\citep{molchanov2016pruning, he2017channel}. \citet{hu2016network} proposed the average percentage of zeros (APoZs) of the activation layer output serving as a pruning criterion. \citet{Liu_2017_ICCV} proposed the scalar parameters representing channels' significance and used the penalty term to push them toward zero during training. 

\subsubsection{Pruning Strategy}
Pruning strategy affects the performance of pruned networks. Iterative pruning follows the pruning-fine-tuning pattern with settings such as the pruning ratio in one iteration, pruning interval, and number of iterations. \citet{zhu2017prune} proposed a gradual sparse ratio schedule with a unified pruning interval for unstructured pruning and was followed by subsequent studies~\citep{gale2019state, liu2021sparse}. \citet{molchanov2016pruning, molchanov2019importance,ding2021resrep} implemented the iterative pruning on channel level. \citet{frankle2018lottery} proposed the lottery hypothesis with a fixed pruning ratio and a longer pruning interval while inheriting the surviving parameters of the initial model after each iteration. \citet{kim2022exploring} proved the lottery hypothesis in SNN pruning. One-shot pruning refers to achieving the target sparsity once. \citet{you2019drawing} proposed using the Hamming distance of pruning masks in the early stages of training to determine the pruning time. \citet{kim2022exploring} implemented the above method on SNN pruning.

\section{Preliminaries}
\subsection{Spiking Neural Network}
In SNNs, a neuron receives signals from model inputs or pre-synaptic neurons and modulates the membrane potential. When the membrane potential exceeds the threshold, An output spike is generated and sent to the post-synaptic neuron. The fire process is described as:
\begin{subequations}
\begin{align}
s[t] = \Theta(h[t]-V_{threshold}), \label{eq10a}
\\
\Theta(x) = 0, x < 0~ otherwise~ 1, \label{eq10b}
\end{align}    
\end{subequations}
where $t$ represents the current time step and $h[t]$  represents current membrane potential. $V_{threshold}$ is the threshold voltage. The neuron model formalizes the processes of charging and resetting of neurons. Following previous SNN works, we use the Leaky Integrate and Fire (LIF) model with a simple form and robust performance.

We employ spatio-temporal backpropagation (STBP)~\citep{wu2018spatio} in the learning process. To address the non-differentiability issue of Eq.~\ref{eq10b} at zero, a surrogate function~\citep{wu2018spatio, wu2019direct, zheng2021going} is commonly utilized during the backward phase to replace Eq.~\ref{eq10b}. We use
\begin{subequations}
\begin{align}
g(x) &= \frac{1}{\pi} arctan(\pi x) + \frac{1}{2},  \label{eq11a}
\\
g'(x) &= \frac{1}{1+\pi^2 x^2}. \label{eq11b}
\end{align}    
\end{subequations}
Eq.~\ref{eq11a} defines the surrogate function utilized in this study. And the derivative, Eq.~\ref{eq11b}, is employed to approximate the gradient in the backward phase.

\subsection{Criticality in Neuroscience}
Neuroscience suggests that the critical state plays a vital role in the brain's efficient information processing~\citep{beggs2012being, di2018landau}. This theory, known as the critical brain hypothesis~\citep{turing2009computing}, asserts that the brain operates in a critical state. In this state, the brain is highly sensitive to any input that can alter its activity. Even minimal stimuli can trigger a rapid cascade of neuronal excitation, facilitating information transmission throughout the brain and improving complex feature extraction~\citep{kinouchi2006optimal, beggs2008criticality, shew2009neuronal, beggs2012being}.

Previous research has observed the criticality in neural systems at different scales~\cite {hesse2014self, heiney2021criticality}. Self-organized criticality (SOC) refers to the ability of a dynamic system to tune itself toward the critical state effectively. \citet{herz1995earthquake} first indicated a mathematical equivalence between SOC models and LIF neuron networks. \citet{gal2013self} established the association between SOC and the single neuron according to the experimental phenomenon. They found that the cortical neurons of newborn rats keep around a barely-excitable state, exhibiting characteristics of SOC. Activities push the neuron towards an excitatory state, while regulatory feedback pulls it back at a longer time scale.

\section{Method}
\subsection{Criticality from Biology to SNNs}
We aim to exploit criticality in SNN pruning to improve feature extraction capability in pruned models and accelerate the entire pruning process. To analyze criticality in SNNs, we employ information entropy and correlate criticality with the maximization of feature information entropy.

\subsubsection{Criticality of Single Neuron} \label{sec:singleC}
We start from a single neuron to analyze the criticality in SNNs since it is the the fundamental unit exhibiting criticality in neuroscience~\citep{gal2013self}. \citet{gal2013self} reveal that criticality is correlated with the barely-excitable state of neurons. \citet{plesser2000noise, maass2014noise} indicate that the magnitude of the membrane potential is positively correlated with the excitation probability. As the membrane potential crosses the threshold voltage, the firing probability rapidly increases from 0 to 1. Following this line, we propose that neuronal criticality is related to the distance between the membrane potential and the threshold voltage. The behavior of a neuron is considered to have higher criticality when the membrane potential is closer to the threshold voltage. 

To further explore the relationship between criticality and neuronal dynamics, information entropy is employed to quantify the information transmission capacity of neurons and SNNs. We compute and maximize the information entropy of the neuron's output. The information entropy of the output of a single neuron in one time step is defined as
\begin{equation}
H(X) = -\sum_{x \in \mathcal{X}}{p(X=x)\ln{p(X=x)}}, \label{eqn1}
\end{equation}
where the random variable $X$ represents the neuron's output, with a value set of $\{0, 1\}$. We denote the firing probability $p(X=1)$ as $p_f$, and Eq.~\ref{eqn1} can be reformulated as
\begin{align}
& H(X) = -p_{f}\ln{p_{f}}-(1-p_{f})\ln{(1-p_{f})}. \label{eqn2}
\end{align}
Eq.~\ref{eqn2} is a concave function with respect to $p_f$, achieving its maximum value of $\ln{2}$ when $p_f = 0.5$.

We introduce $p(u)$ to represent the neuron firing probability $p_f$, where $u$ is the membrane potential of the neuron at the current time step, decoupling the firing probabilities across different time steps. As previously mentioned, experiments have demonstrated a rapid increase in firing probability from 0 to 1 as the membrane potential increases. And $p(u)=0.5$ is achieved when the membrane potential is equal to the threshold voltage. Since Eq.~\ref{eqn2} is concave, the information entropy of the neuron's output is greater when the membrane potential is closer to the threshold voltage.

\subsubsection{Criticality of Neuron Population}

For populations of neurons, we define the joint probability distribution of the output feature map of a layer as $p(\mathbf{X})$. The information entropy of the feature map output can be represented as
\begin{equation}
H(\mathbf{X}) = -\sum_{\mathbf{x} \in \mathbf{\mathcal{X}}}p(\mathbf{X}=\mathbf{x}) \ln{p(\mathbf{X}=\mathbf{x})}, \label{eqn4}
\end{equation}
where $\mathbf{X}$ is a random vector representing the values of feature maps. As the size of the feature maps increases, the sample space of $\mathbf{X}$ grows exponentially, making it difficult to directly analyze the maximum value of $H(\mathbf{X})$. In fact, we estimate the joint entropy $H(\mathbf{X})$ using the sum of the individual neuron entropies $\sum_{i=1}^{n}{H(X_i)}$. We have
\begin{align}
& \sum_{i=1}^{n} H(X_i) - H(\mathbf{X}) \label{eqn5} \\ 
& = \sum_{\mathbf{x} \in \mathbf{\mathcal{X}}} p(\mathbf{x}) \ln p(\mathbf{x}) - \sum_{i=1}^{n} \sum_{a \in \mathcal{X}_i} p(X_i=a) \ln p(X_i=a) \label{eqn6} \\
& = \sum_{\mathbf{x} \in \mathbf{\mathcal{X}}} p(\mathbf{x}) \ln p(\mathbf{x}) - \sum_{i=1}^{n} \sum_{a \in \mathcal{X}_i} \sum_{\mathbf{x} \in \mathbf{\mathcal{X}}} p(\mathbf{x}) I_{x_i=a}(\mathbf{x}) \ln p(X_i=a) \label{eqn7} \\
& = \sum_{\mathbf{x} \in \mathbf{\mathcal{X}}} p(\mathbf{x}) \ln p(\mathbf{x}) - \sum_{\mathbf{x} \in \mathbf{\mathcal{X}}} p(\mathbf{x}) \ln{\prod_{i=1}^{n}p(X_i=x_i)} \label{eqn8} \\
& = \sum_{\mathbf{x} \in \mathbf{\mathcal{X}}} p(\mathbf{x}) \ln{\frac{p(\mathbf{x})}{\prod_{i=1}^{n}p(X_i=x_i)}} . \label{eqn9}
\end{align}
In Eq.~\ref{eqn7}, we expand $p(X_i=a)$ with $\mathbf{x} \in \mathbf{\mathcal{X}}$. $I_{x_i=a}(\mathbf{x})$ is an indicator function, which is 1 only when $x_i=a$ and 0 otherwise. The difference between the full entropy and the sum of individual neuron entropies is transformed into Eq.~\ref{eqn9}. The inequality follows from the fact that Eq.~\ref{eqn9} represents the Kullback-Leibler (KL) divergence between the probability distributions $p(\mathbf{x})$ and $\prod_{i=1}^{n}p(X_i=x_i)$.The KL divergence indicates the distance between $H(\mathbf{X})$ and $\sum_{i=1}^{n}{H(X_i)}$ and the distance is 0 when $p(\mathbf{x}) = \prod_{i=1}^{n}p(X_i=x_i)$.

We similarly introduce $p(u_i)$ to represent $p(X_i=1)$, where $u_i$ is the membrane potential of the $i$-th neuron at the current time step. The distribution of membrane potentials is reorganized through the Batch Normalization (BN) during the forward propagation process, reducing the correlation of firing rate $p(u_i)$ between neurons and making $p(\mathbf{x})$ close to $\prod_{i=1}^{n}p(X_i=x_i)$. On this basis, we estimate $H(\mathbf{X})$ using $\sum_{i=1}^{n} H(X_i)$, which has been proven to be reasonable during the maximization process~\citep{dayan2005theoretical}. From Eq.~\ref{eqn2}, it is easy to obtain that $\sum_{i=1}^{n} H(X_i)$ is concave in $R^n$, and it takes the maximum value when $p(u_i)=0.5, \forall i$. As mentioned in Section \ref{sec:singleC}, $p(u_i)$ rapidly increases from 0 to 1 when $u_i$ crosses the threshold voltage. $p(u_i)$ reaches $0.5$ at a point where $u_i$ is infinitely close to the threshold voltage. In this case, the neuron approaches a critical state, transmitting a greater amount of information at both the single neuron and feature map levels.

\subsection{Metric of Neuron Criticality}

So far, we have known that when the membrane potential is closer to the threshold voltage, the neuron's criticality is higher and the output feature carries more information. To integrate criticality into SNN pruning, it is necessary to identify an appropriate metric to assess the criticality of neurons. The primary challenge lies in capturing the rapid changes of criticality near the threshold voltage and ensuring consistent, low-cost computation during model training.

We propose that the derivative of the surrogate function, such as  $g'$ in Eq.~\ref{eq11b}, can serve as a criticality metric. It offers several advantages: (1) The derivatives of widely-used surrogate functions reach the peak when the membrane potential equals the threshold voltage and rapidly decreases with increasing distance, effectively reflecting the criticality changes of a neuron's behavior. (2) The derivative of the surrogate function can be directly obtained during model training without additional computational cost. (3) The derivative of the surrogate function has a unified value range, enabling global ranking.

\begin{figure*}[htb]
	\begin{subfigure}{0.51\linewidth}
		\centering
	\includegraphics[width=1.0\linewidth]{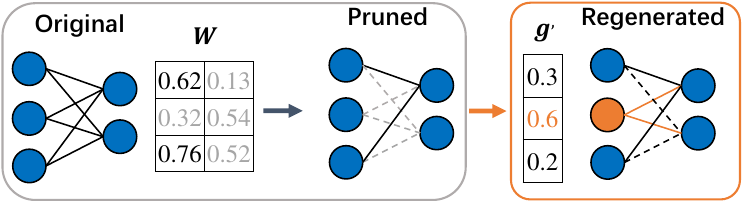}
		\caption{Unstructured Pruning}
		\label{fw}
	\end{subfigure}
        \centering
	\begin{subfigure}{0.48\linewidth}
		\centering
	\includegraphics[width=1.0\linewidth]{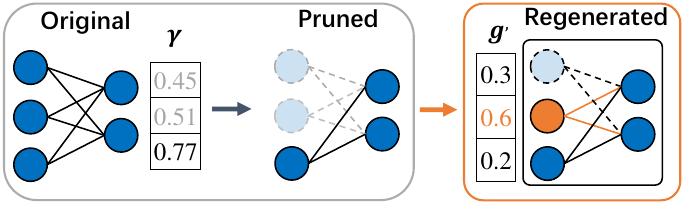}
		\caption{Structured Pruning}
		\label{fc}
	\end{subfigure}
        \centering

	\caption{Schematic view of Pruning-Regeneration Process. Faded structures and connections with dotted lines are pruned. The orange ones with higher criticality scores are regenerated in our method. (a) Unstructured pruning via global magnitude pruning criterion and regeneration based on criticality. (b) Structured pruning via the scale factor $\gamma$ (BN layer parameter) and regeneration based on criticality.}
	\label{framework}
\end{figure*}

\subsection{Overall Pruning Strategy} \label{sec:method}

\subsubsection{Regeneration via Criticality}

Our goal is to develop a mechanism compatible with exist pruning strategies to leverage both the criticality and the established advantages of SOTA methods. To achieve this, we combine the criticality metric mentioned above with a neuron regeneration mechanism. The regeneration mechanism, inspired by neuroregeneration in the nervous system where new neurons and connections are synthesized to recover damage, allows the pruned network to regenerate lost connections, thereby improving performance after pruning.

We implement a regeneration mechanism based on neuron criticality to maintain the critical state of the pruned model. Specifically, to avoid the sparsity of the model after regeneration from being lower than the target, we first temporarily expand the target sparsity at the current pruning iteration $s_t$ to $s_t'$:
\begin{equation}
s_t’ = s_t + r (1-s_t), \label{eq5} 
\end{equation}
where $t$ represents current pruning iteration and $r$ is the regeneration rate. $s_t$ is the target sparsity, and $s_t'$ is the temp sparsity for pruning before regeneration at the current iteration. After each pruning iteration, we re-sort all pruned structures (neurons or synapses) in the pruned model based on the criticality scores of the corresponding neurons and reactivate the top k with the highest scores:
\begin{equation}
S_{\text{new}} = S + \text{TopK}(C(S')), \label{eq1}
\end{equation}
where $S$ is the set of surviving pruning structures after the pruning iteration. $S’$ is the set of all pruned structures in current iteration. $C(x)$ denotes the criticality score of $x$. To compute the criticality score, we have
\begin{subequations}
\begin{align}
s(e) & = \text{aggregate}\left(\frac{1}{T} \sum^{T}_{i=0}{g'(u_{e,i})}\right), \label{eq2a} \\
C(e) & = \frac{1}{N} \sum_{i=1}^{N}{s_i(e)}. \label{eq2b}
\end{align}
\label{eq2}
\end{subequations}
Eq.~\ref{eq2a} and \ref{eq2b} illustrate the approach to obtaining criticality scores for a neuron $e$. $u_{e,i}$ represents the membrane potential of $e$ at $i$th time step. $T$ is the total number of time steps. $g'$ is the derivative of the surrogate function. $N$ is the total number of samples used to compute the criticality score. $aggregate$ represents the way to aggregate the results for the corresponding pruning structure. 

For the fully connected layer, we directly compute the average criticality score for each neuron. For convolutional layers, we explore two approaches: mean aggregation and max aggregation. We finally employ max operation due to its superior performance in the experiments. We consider that max aggregation preserves the highest response within the corresponding region, preventing robust responses from being diminished by neighboring elements within the feature map.

\subsubsection{Unstructured Pruning}
So far, we have known how to calculate the criticality scores of neurons. we perform unstructured (connection) pruning on SNNs as depicted in Fig. \ref{fw}. Our approach employs iterative magnitude pruning and incorporates a regeneration mechanism based on neuron criticality. 
We choose the simple global magnitude pruning criterion $|w|$ to achieve connection level sparsity in deep SNNs. Specifically, we have:
\begin{equation}
s(w) = |w|, \label{aeq1}
\end{equation}
where $w$ is the connection weight and $s(w)$ is the importance score of the connection.

\begin{algorithm}[htb]
\caption{The pseudo code of unstructured pruning}
\label{alg:a1}
\begin{algorithmic}[1] 
\Require target sparsity $s_f$, pruning interval $\Delta t$, total period $T_f$ and regeneration ratio r. 
\Param Model weights $W$.
    \State Let $n \leftarrow 0$.
    \For{each training step $t$}
    \State TrainOneStep()
        \If {$t < T_f$ and $(t \bmod \Delta t) == 0$}
            \State $n=n+1$
            \State $s_t\leftarrow $ CurrentSparsity($s_f, n, T_f, \Delta t$) \Comment{Eq.~\ref{aeq2}} 
            \State $s'_{t} \leftarrow$ ExtendSparsity($s_t, r$) \Comment{Eq.~\ref{eq5}}
            \State $W = $ Pruning($|W|, s'_{t}$) \Comment{Eq.~\ref{aeq1}}
            \State $C(W) \leftarrow $ Criticality($(u_{currentStep}$) \Comment{Eq.~\ref{eq2}}
            \State $W =$ Regeneration($C(W), s'_{t}-s_t$)  \Comment{Eq.~\ref{eq1}}
        \EndIf
    \EndFor
\end{algorithmic}
\end{algorithm}

To best maintain the criticality of the pruned model, we employ a gradual pruning scheme and iteratively prune the network until reaching the final target sparsity. The gradual pruning scheme is
\begin{equation}
s_t = s_f - s_f(1- \frac{n\Delta t}{T_f})^3, \label{aeq2} 
\end{equation}
where $s_f$ represents the final target sparsity. $n$ is the current pruning iteration. $T_f$ is the total period of pruning. $\Delta t$ is a constant and represents the interval of pruning iterations. 

We present the overall pruning strategy with regeneration in Algorithm \ref{alg:a1}. The regeneration based on neuron criticality is included in each pruning iteration. Specifically, we collect the values of $g'(u)$ from the last training iteration before pruning and calculate the average criticality score. Then, we perform the global regeneration after pruning.

\subsubsection{Structured Pruning}

To evaluate the effectiveness of applying criticality in once pruning, we conduct one-shot structured pruning as depicted in Fig.~\ref{fc}. The overall pruning strategy is presented in Algorithm \ref{alg:a2}. Specifically, we leverage the parameter $\gamma$ in batch normalization layers as the pruning criterion \citep{Liu_2017_ICCV} to achieve structured channel-level sparsity, and incorporate L1 sparsity regularization during training:
\begin{equation}
L = \sum_{(x,y)}{l(f(x,W),y)} +\lambda \sum_{\gamma \in \Gamma}{|\gamma|}, \label{aeq3} 
\end{equation}
where $(x,y)$ is the input and target pair. $W$ is network weights. The first sum term denotes train loss in the backward process. The second sum term is the L1 sparsity regularization. After training, we prune channels according to the magnitude of $\gamma$. Then, we calculate the criticality score for each channel by collecting the average values of the surrogate function derivative from the entire training dataset. Finally, we execute regeneration with criticality scores and fine-tune the pruned network.

\begin{algorithm}[t]
\caption{The pseudo code of structured pruning}
\label{alg:a2}
\begin{algorithmic}[1] 
\Require target sparsity $s_t$, and regeneration ratio r.
\Param Model channels $chs$ and scaling factor set $\Gamma$. 
\textbf{Train process:}
    \State Train model with L1 sparsity regularization. \Comment{Ep.\ref{aeq3}}
\textbf{Pruning and Regeneration:}
    \State $s'_{t} \leftarrow $ ExtendSparsity($s_t, r$) \Comment{Eq.~\ref{eq5}}
    \State $chs = $ Pruning($chs, s'_t, |\Gamma|$) \Comment{pruning by $\gamma$}
    \State $C(chs) \leftarrow $ Criticality($(u_{trainSet}$) \Comment{Eq.~\ref{eq2}}
    \State $chs =$ Regeneration($C(chs), s'_{t}-s_t$)  \Comment{Eq.~\ref{eq1}}
\textbf{Fine-tuning process:}
    \State Fine-tuning model without L1 sparsity regularization.
\end{algorithmic}
\end{algorithm}

\section{Experiment}

\subsection{Experimental Setup}
We evaluate our method on a shallow SNN(6-conv-2-fc) and two deep SNNs VGG16~\citep{simonyan2014very} and ResNet19~\citep{he2016deep} for three image classification datasets CIFAR10, CIFAR100~\citep{krizhevsky2009learning} and Tiny-ImageNet~\citep{hansen2015tiny}. All experiments are executed on an RTX 3090 GPU and the implementation is based on PyTorch. The networks are trained using SGD optimizer with momentum 0.9, learning rate 0.3 and  batch size 128. We set 200 epochs for unstructured pruning for CIFAR10 and CIFAR100 (100 for tiny-ImageNet) and 160 epochs for structured pruning. To be consistent with previous works, the time step is set to 5 for all experiments, and batch normalization (BN) is used with the first convolutional layer serving as the encoder to transform input images into spikes. A step lrscheduler is used for structured pruning, while a cosine lrscheduler is applied for unstructured pruning. The detail settings are illustrated in Table \ref{tab:commonE}.

\begin{table}[htb]
\footnotesize
\caption{The detailed settings of hyperparameters for all experiments.}
\centering
\begin{threeparttable}
\begin{tabularx}{0.46\textwidth}{
   >{\centering\arraybackslash}X 
   >{\centering\arraybackslash}X 
   >{\centering\arraybackslash}X }
\toprule
\textbf{Parameter} & \textbf{Description}  & \textbf{Value} \\
\midrule
\multicolumn{3}{c}{\textbf{Global Setting}} \\
\midrule
$T$ & \# Timestep & 5 \\
$\tau$ & Membrane constant & $4/3$ \\
$V_{\text{threshold}}$ & Threshold voltage & 1.0 \\
$V_{\text{reset}}$ & Reset voltage & 0.0 \\
\midrule
\multicolumn{3}{c}{\textbf{For Unstructured Pruning}} \\
\midrule
$wd$ & Weight decay & 1e-4 \\
$\Delta t$ & Pruning interval & 2000 \\
$s_f$ & Final sparsity & [0.90, 0.95, 0.98] \\
$r$ & Regeneration ratio & [0.2, 0.1, 0.01] \\
\midrule
\multicolumn{3}{c}{\textbf{For Structured Pruning}} \\
\midrule
$N_1, N_2$ & Drop (10x) epochs & [80, 120] \\
$\gamma$ & Scaling factor of L1 & 1e-4 \\
$percent^1$ & Target sparsity & [0.512, 0.658] \\
$r$ & Regeneration ratio & [0.3, 0.1] \\
\bottomrule
\end{tabularx}
\begin{tablenotes}
\footnotesize
\item[1] Choosing special values to achieve the same flops reduction as SOTA for fair comparison.
\end{tablenotes}
\end{threeparttable}
\label{tab:commonE}
\end{table}

\subsection{Accuracy and Efficiency Evaluations}

\begin{table*}[!h]
    \centering
    \footnotesize
    \begin{threeparttable}
    \caption{Performance comparison between our method and previous works for unstructured pruning. The results reported with (mean$\pm$std) are run with three random seeds. We mark the best results in bold. }
    \label{tab:wc}
    \begin{tabular}{c c c c c c c}
        \toprule
            \textbf{Pruning Method} & \textbf{Dataset} & \textbf{Arch.} & \textbf{Base Acc. (\%)} & \textbf{\#Epoch} & \textbf{Pruned Acc. (\%)} & \textbf{Sparsity (\%)} \\
        \midrule
            \multirow{2}{*}{Grad R~\citep{ijcai2021-236}} & 
            \multirow{2}{*}{CIFAR10} & 
            \multirow{2}{*}{6Conv, 2Fc}  & 
            \multirow{2}{*}{92.84} & \multirow{2}{*}{2048} & 91.47 & 97.65  \\ 
            & & & & & 89.32 & 99.27  \\
        \midrule
            \multirow{2}{*}{STDS~\citep{chen2022state}} & 
            \multirow{2}{*}{CIFAR10} & 
            \multirow{2}{*}{6Conv, 2Fc}  & 
            \multirow{2}{*}{92.84} & \multirow{2}{*}{-} & 92.49 & 97.77  \\ 
            & & & & & 90.21 & 99.25  \\
       \midrule
            \multirow{2}{*}{RCMO-SNN~\citep{chen2023resource}} & 
            \multirow{2}{*}{CIFAR10} & 
            \multirow{2}{*}{6Conv, 2Fc}  & 
            \multirow{2}{*}{92.88} & \multirow{2}{*}{2048} & 92.75 & 97.71  \\ 
            & & & & & 90.32 & 99.31  \\
        \midrule
            \multirow{2}{*}{Ours} & 
            \multirow{2}{*}{CIFAR10} & 
            \multirow{2}{*}{6Conv, 2Fc}  & 
            \multirow{2}{*}{92.79} & \multirow{2}{*}{\textbf{200}} & \textbf{92.77$\pm$0.21} & 97.71  \\ 
            & & & & & \textbf{90.45$\pm$0.16} & 99.31  \\
        \midrule

        \midrule
            \multirow{3}{*}{IMP~\citep{kim2022exploring}} & 
            \multirow{3}{*}{CIFAR100} & 
            \multirow{3}{*}{VGG16}  & 
            \multirow{3}{*}{69.86} & 2260 & 68.90 & 89.91  \\ 
            & & & & 3380 & 68.00 & 95.69 \\
            & & & & 4220 & 66.02 & 98.13 \\
        \midrule
            \multirow{3}{*}{NDSNN~\citep{dac2023}} & 
            \multirow{3}{*}{CIFAR100} & 
            \multirow{3}{*}{VGG16}  & 
            \multirow{3}{*}{69.86} & \multirow{3}{*}{300} & 68.07 & 90.00  \\ 
            & & & &  & 66.73 & 95.00 \\
            & & & & & 63.51 & 98.00 \\
        \midrule
            \multirow{3}{*}{Ours} & 
            \multirow{3}{*}{CIFAR100} & 
            \multirow{3}{*}{VGG16}  & 
            \multirow{3}{*}{72.89} & \multirow{3}{*}{\textbf{200}} & \textbf{72.69$\pm$0.15} & \textbf{90.00}  \\ 
            & & & & & \textbf{72.23$\pm$0.19} & \textbf{95.69} \\
            & & & & & \textbf{70.70$\pm$0.20} & \textbf{98.13} \\
        \midrule
        \midrule
            \multirow{2}{*}{Grad R~\citep{ijcai2021-236}} & 
            \multirow{2}{*}{CIFAR100} & 
            \multirow{2}{*}{ResNet19}  & 
            \multirow{2}{*}{71.34} & \multirow{2}{*}{2048} & 67.47 & 94.92  \\ 
            & & & & & 67.31 & 97.65  \\
        \midrule
            \multirow{3}{*}{IMP ~\citep{kim2022exploring}} & 
            \multirow{3}{*}{CIFAR100} & 
            \multirow{3}{*}{ResNet19}  & 
            \multirow{3}{*}{71.34} & 2860 & 71.38 & 89.91  \\ 
            & & & & 3380 & 70.54 & 95.69  \\ 
            & & & & 4220 & 67.35 & 98.13  \\
        \midrule
            \multirow{2}{*}{RCMO-SNN~\citep{chen2023resource}} & 
            \multirow{2}{*}{CIFAR100} & 
            \multirow{2}{*}{ResNet19}  & 
            \multirow{2}{*}{74.71} & 3100 & 72.67 & 95.19  \\ 
            & & & & 3940 & 70.80 & 97.31  \\     
        \midrule
        \multirow{3}{*}{NDSNN~\citep{dac2023}} & 
            \multirow{3}{*}{CIFAR100} & 
            \multirow{3}{*}{ResNet19}  & 
            \multirow{3}{*}{71.94} & \multirow{3}{*}{300} & 70.08 & 90.00  \\ 
            & & & &  & 68.95 & 95.00 \\
            & & & & & 65.48 & 98.00 \\
        \midrule
            \multirow{3}{*}{\textbf{Ours}} & 
            \multirow{3}{*}{CIFAR100} & 
            \multirow{3}{*}{ResNet19}  & 
            \multirow{3}{*}{74.31} & \multirow{3}{*}{\textbf{200}} & \textbf{73.65$\pm$0.23} & \textbf{90.00}  \\ 
            & & & & & \textbf{73.01$\pm$0.06} & \textbf{95.19}  \\ 
            & & & & & \textbf{71.48$\pm$0.12} & \textbf{98.13}  \\ 
        \midrule
        \midrule
            \multirow{3}{*}{RigL~\citep{evci2020rigging}} & 
            \multirow{3}{*}{Tiny-ImageNet} & 
            \multirow{3}{*}{ResNet19}  & 
            \multirow{3}{*}{50.32} & \multirow{3}{*}{300} & 49.49 & 90.00  \\ 
            & & & & & 40.40 & 95.00  \\ 
            & & & & & 37.98 & 98.00  \\
        \midrule
            \multirow{3}{*}{NDSNN~\citep{dac2023}} & 
            \multirow{3}{*}{Tiny-ImageNet} & 
            \multirow{3}{*}{ResNet19}  & 
            \multirow{3}{*}{50.32} & \multirow{3}{*}{300} & 49.25 & 90.00  \\ 
            & & & & & 47.45 & 95.00  \\ 
            & & & & & 45.09 & 98.00  \\
        \midrule
            \multirow{3}{*}{IMP \tnote{1} ~\citep{kim2023exploring}} & 
            \multirow{3}{*}{Tiny-ImageNet} & 
            \multirow{3}{*}{ResNet19}  & 
            \multirow{3}{*}{58.00} & \multirow{3}{*}{100} & 54.94  & 90.00  \\ 
            & & & & & 54.54 & 95.00  \\ 
            & & & & & 51.03 & 98.00  \\
        \midrule
            \multirow{3}{*}{\textbf{Ours}} & 
            \multirow{3}{*}{Tiny-ImageNet} & 
            \multirow{3}{*}{ResNet19}  & 
            \multirow{3}{*}{59.85} & \multirow{3}{*}{\textbf{100}} & \textbf{58.00$\pm$0.16} & \textbf{90.00}  \\ 
            & & & & & \textbf{57.95$\pm$0.25} & \textbf{95.00}  \\ 
            & & & & & \textbf{56.24$\pm$0.09} & \textbf{98.00}  \\   
        \bottomrule
    \end{tabular}
    \begin{tablenotes}
    \footnotesize
    \item[1] We run the code from~\citep{kim2022exploring} to obtain results of Tiny-ImageNet.
    \end{tablenotes}
    \end{threeparttable}
\end{table*}

\begin{table}[htb]
\footnotesize
\caption{Performance comparison between our method to~\citep{Liu_2017_ICCV} and~\citep{wang2021neural}. We report the test accuracy of VGG16 and ResNet19 models after structured pruning on CIFAR100 and mark the best results in bold.}

\centering
\begin{tabularx}{0.48\textwidth}{l 
   >{\centering\arraybackslash}X 
   >{\centering\arraybackslash}X 
  }
\toprule

\textbf{Pruning Method} & \textbf{Pruned Acc. (\%)} & \textbf{FLOPs $\downarrow$ (\%)} \\

\midrule
\midrule
\multicolumn{3}{c}{VGG16} \\
\midrule
\multirow{2}{*}{NSlim~\citep{Liu_2017_ICCV}}  & 71.93 & 43.57 \\
 & 70.41 & 57.47 \\
\midrule
\multirow{2}{*}{Greg~\citep{wang2021neural}} & 72.05 & 43.65 \\
& 70.13  & 57.21 \\
\midrule
\multirow{2}{*}{\textbf{Ours}} & \textbf{72.65} & \textbf{43.63} \\ 
& \textbf{71.47} & \textbf{57.70} \\

\midrule
\midrule
\multicolumn{3}{c}{ResNet19} \\
\midrule
\multirow{2}{*}{NSlim~\citep{Liu_2017_ICCV}}  & 72.98 & 48.05 \\
& 70.13 & 66.92 \\
\midrule
\multirow{2}{*}{Greg~\citep{wang2021neural}}  & 73.45 & 47.10 \\
& 70.88 & 67.69 \\
\midrule
\multirow{2}{*}{\textbf{Ours}} & \textbf{74.63}  & \textbf{47.26} \\
& \textbf{72.51} & \textbf{67.85} \\
\bottomrule
\end{tabularx}
\label{tab:cc}
\end{table}

\subsubsection{Unstructured Pruning}
We evaluate our method in Table \ref{tab:wc} and compare it with state-of-the-art (SOTA) pruning methods for SNNs, including Grad R~\citep{ijcai2021-236}, STDS~\citep{chen2022state}, RigL~\citep{evci2020rigging}, IMP~\citep{kim2022exploring}, NDSNN~\citep{dac2023}, and RCMO-SNN~\citep{chen2023resource}. We report accuracy and pruning costs (\#Epoch) at different sparsity levels. On CIFAR10 with the shallow 6conv-2fc SNN, our method outperforms previous approaches at all sparsity levels with lower pruning costs. On the more complex datasets CIFAR100 and Tiny-ImageNet, our method exhibits even more significant improvements. As shown in Table \ref{tab:wc}, On CIFAR100, we achieve higher accuracy compared to IMP and RCMO-SNN with a pruning cost reduction ranging from 91.15\% to 95.26\% on VGG16 and ResNet19. For Tiny-ImageNet, our method demonstrates from 3.06\% to 5.21\% performance advantages over IMP with the same pruning costs. Our method also achieves smaller accuracy losses compared to nearly all SOTA methods. While IMP reports a smaller accuracy loss on ResNet19 for CIFAR100, it is based on a lower base accuracy. \citet{chen2023resource} reproduced IMP, achieving a base accuracy that is 3.37\% higher than the reported results in~\citep{kim2023exploring}. Our reproduction confirms this, showing that IMP incurs greater accuracy losses.

\subsubsection{Structured Pruning}
In Table \ref{tab:cc}, we evaluate the performance of our method on CIFAR100 for structured pruning. As the first work in implementing structured pruning on deep SNNs, we compare our method with the SOTA methods for ANN pruning NSlim~\citep{Liu_2017_ICCV} and GReg~\citep{wang2021neural}, reporting the pruned test accuracy and the float point operations reduction (FLOPs $\downarrow$ in Table~\ref{tab:cc}) in inference. Additionally, we take NSlim as our baseline because our method develops based on it. Our method outperforms GReg \citep{wang2021neural}, which is more complex and computationally expensive. It also shows significant improvements compared to the baseline, all while maintaining the equivalent level of FLOPs.

\subsection{Effect of Neuron Criticality}

\begin{figure}[ht]
    \centering
    \begin{subfigure}{0.49\linewidth}
	\centering
	\includegraphics[width=1.0\linewidth]{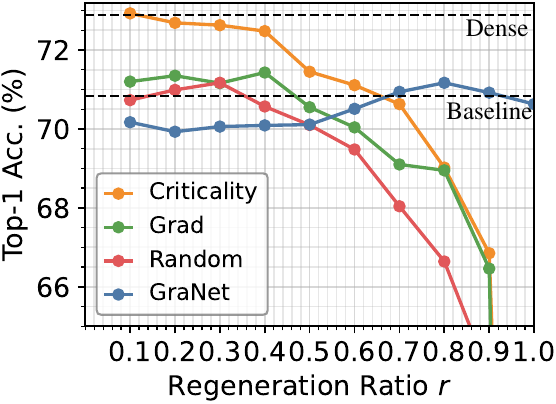}
        \caption{}
        \label{rrv}
    \end{subfigure}
    \centering
    \begin{subfigure}{0.49\linewidth}
	\centering
	\includegraphics[width=1.0\linewidth]{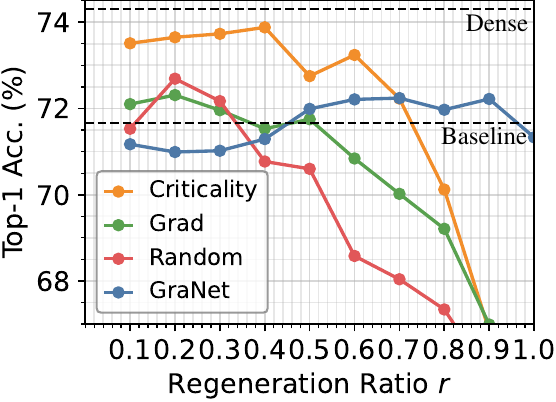}
        \caption{}
        \label{rrc}
    \end{subfigure}
    \caption{Comparison of regeneration methods based on criticality (ours), gradient, random, and GraNet \citep{liu2021sparse}. The two dashed lines represent the performance of the dense model and the baseline, respectively. (a) Accuracy of VGG16 with 90\% sparsity for CIFAR100. (b) Accuracy of ResNet19 with 90\% sparsity for CIFAR100.}
    \label{rr}
\end{figure}

We compare the baseline and pruning based on neuron criticality on CIFAR100. The results of the ablation study are shown in Table \ref{tab:ablation}, where the proposed method outperforms the baseline at all sparsity levels. Furthermore, to evaluate the contribution of the criticality-based regeneration mechanism, we perform a performance analysis in Fig. \ref{rr} using various regeneration metrics. ``Criticality" refers to the method based on criticality. ``Grad" employs parameter gradients at the current iteration. ``Random" adopts random ranking, and ``GraNet" represents the GraNet gradient regeneration strategy~\citep{liu2021sparse}. We observe that criticality-based regeneration consistently outperforms Grad and Random regeneration, highlighting our method's superiority.

\subsection{Effect of Regeneration Ratio}
To evaluate the robustness of the criticality-based regeneration, we conduct performance analysis in Fig. \ref{rr} using different metrics with different regeneration ratios $r$. The two dashed lines represent the dense model and baseline performance, respectively. We observe that the criticality-based method exhibits stable performance within the ratio range from 0.1 to 0.6 and higher accuracy than the baseline and other methods, indicative of the robustness over the wide range of $r$. GraNet exhibits different performance dynamics compared to other methods. We attribute this to the cosine decay of the regeneration ratio, gradually decreasing the impact of GMP as training progresses.

\begin{table}[ht]
\footnotesize
\centering
\caption{Test accuracy of VGG16 model with different sparsity on CIFAR100 after unstructured pruning using the baseline and proposed method.}

\begin{tabularx}{0.48\textwidth}{l 
   >{\centering\arraybackslash}X 
   >{\centering\arraybackslash}X 
   >{\centering\arraybackslash}X 
  }
\toprule
Sparsity & 90.00\%  & 95.69\% & 98.13\% \\
\midrule
baseline & 70.84 & 70.56 & 69.98 \\
w/neuron criticality & 72.69 & 72.23 & 70.70 \\
\bottomrule
\end{tabularx}

\label{tab:ablation}
\end{table}

\begin{figure}[htb]
 	\centering
	\begin{subfigure}{0.49\linewidth}
		\centering
		\includegraphics[width=1.0\linewidth]{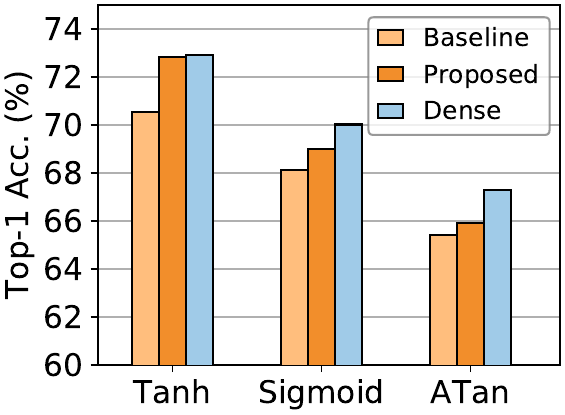}
		\caption{}
		\label{sf}
	\end{subfigure} 	
        \centering
	\begin{subfigure}{0.49\linewidth}
		\centering
		\includegraphics[width=1.0\linewidth]{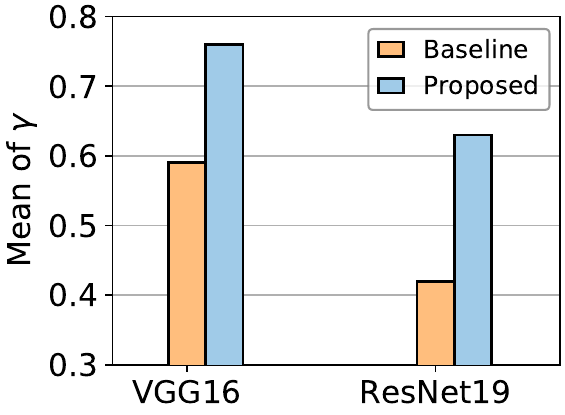}
		\caption{}
		\label{it}
	\end{subfigure}
	\caption{(a) Performance of dense model, the baseline, and the proposed method under different surrogate functions on VGG16 for CIFAR100. (b) Mean importance of non-overlapping channels between our method and the baseline for structured pruning after fine-tuning.}
	\label{re_dif}
\end{figure} 

\subsection{Effect of Surrogate Function} \label{sec:sf}

To evaluate the effectiveness and robustness of the criticality mechanism for different surrogate functions, we report the pruning performance of different surrogate functions on VGG16 for CIFAR100. As shown in Fig. \ref{sf}, we compare three surrogate functions, tanh, sigmoid, and arc tangent, for the dense model, baseline, and proposed method. The proposed method achieves higher performance compared to the baseline across different surrogate functions. The formulas of tanh, sigmoid, and arc tangent functions are respectively represented as  Eq.~\ref{eqa1a}, Eq.~\ref{eqa2a}, Eq.~\ref{eq11a}.

\begin{equation}
g_1(x) = \frac{1}{2} \cdot \frac{tanh(3x)}{tanh(1.5)} + \frac{1}{2},   \label{eqa1a}
\end{equation}

\begin{equation}
g_2(x) = \text{sigmoid}(4x) = \frac{1}{1+e^{-4x}},  \label{eqa2a}
\end{equation}

\section{Discussion}

In order to explore the underlying impact of neuron criticality during pruning and fine-tuning, we present the experimental results and statistical findings for the pruned model in this section for more in-depth analysis.

\subsection{Importance Transition} 
We compare the importance transition in non-overlapping surviving structures between the models obtained by our method and the baseline. To better distinguish non-overlapping structures, we focus on the results of structured pruning. In Fig. \ref{it}, we present the means of $\gamma$ (normalized) of non-overlapping channels in the models through the fine-tuning process with L1 sparsity regularization for two methods. The results show that the channels regenerated by our method exhibit significantly higher importance than those without regeneration. Considering that the regenerated channels initially corresponded to lower $\gamma$ values and higher criticality, this result suggests the following role of our method: identifying structures with potential, enabling the high-criticality structures to occupy key positions and efficiently leading to higher performance gains.

\begin{figure*}[!t]
  	\centering
	\begin{subfigure}{0.245\linewidth}
		\centering
		\includegraphics[width=1.0\linewidth]{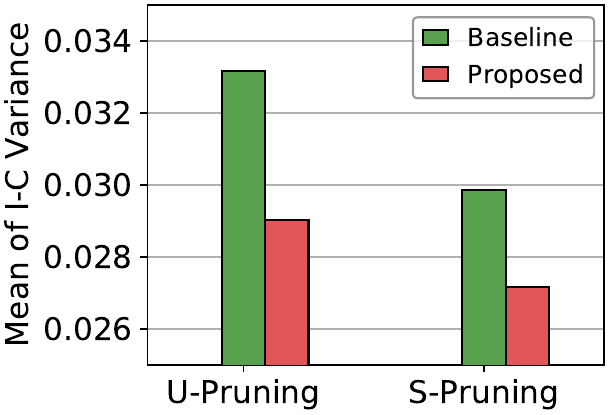}
		\caption{}
		\label{ic}
	\end{subfigure}
  	\centering
	\begin{subfigure}{0.265\linewidth}
		\centering
		\includegraphics[width=1.0\linewidth]{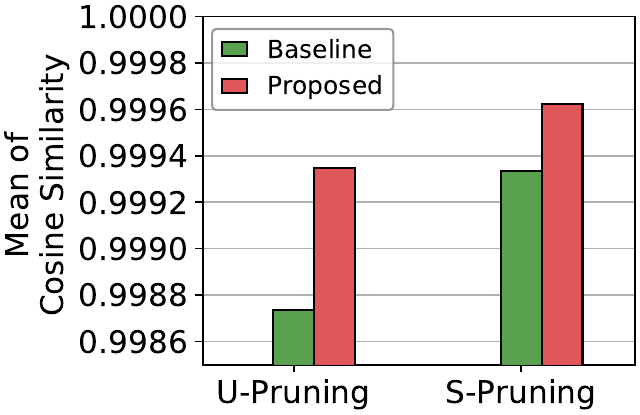}
		\caption{}
		\label{cs}
	\end{subfigure}
  	\centering
	\begin{subfigure}{0.235\linewidth}
		\centering
		\includegraphics[width=1.0\linewidth]{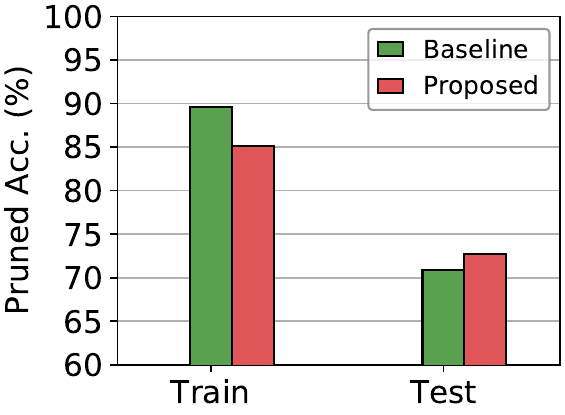}
		\caption{}
		\label{accw}
	\end{subfigure}
   	\centering
	\begin{subfigure}{0.235\linewidth}
		\centering
		\includegraphics[width=1.0\linewidth]{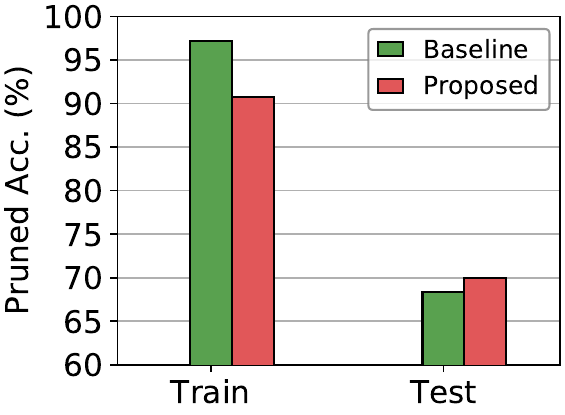}
		\caption{}
		\label{accc}
	\end{subfigure}
	\caption{Comparison of the difference in feature extraction of models pruned by our method and the baseline. Our method achieves more uniform feature representations. (a) Comparison of mean of intra-cluster variance between proposed method and the baseline for VGG16 after unstructured pruning (U-Pruning) and ResNet19 after structured pruning (S-Pruning). (b) Average cosine similarity between the class feature map between the training and test dataset for VGG16 after unstructured pruning (U-Pruning) and ResNet19 after structured pruning (S-Pruning). We compare the results between the proposed method and the baseline. (c) Train accuracy and test accuracy on VGG16 for CIFAR100 after unstructured pruning using the proposed method and the baseline. (d) Train accuracy and test accuracy on ResNet19 for CIFAR100 after structured pruning using the proposed method and the baseline.}
	\label{fe}
\end{figure*}

\begin{figure*}[htb]
        \centering
	\begin{subfigure}{0.245\linewidth}
		\centering
		\includegraphics[width=1.0\linewidth]{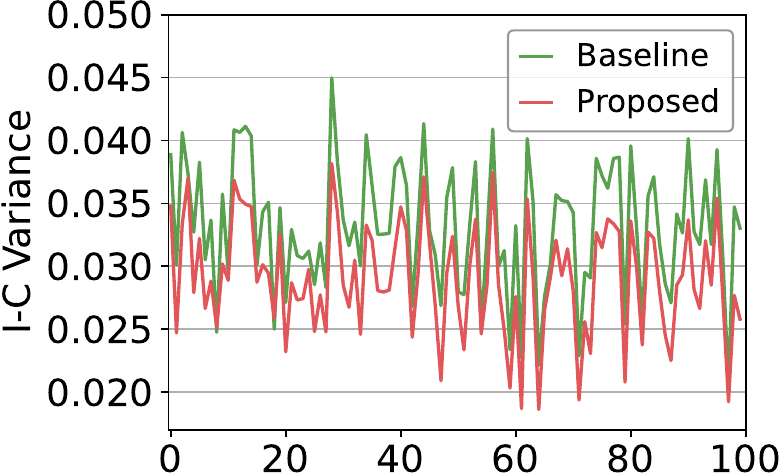}
		\caption{}
		\label{aicw}
	\end{subfigure}
 	\centering
	\begin{subfigure}{0.245\linewidth}
		\centering
		\includegraphics[width=1.0\linewidth]{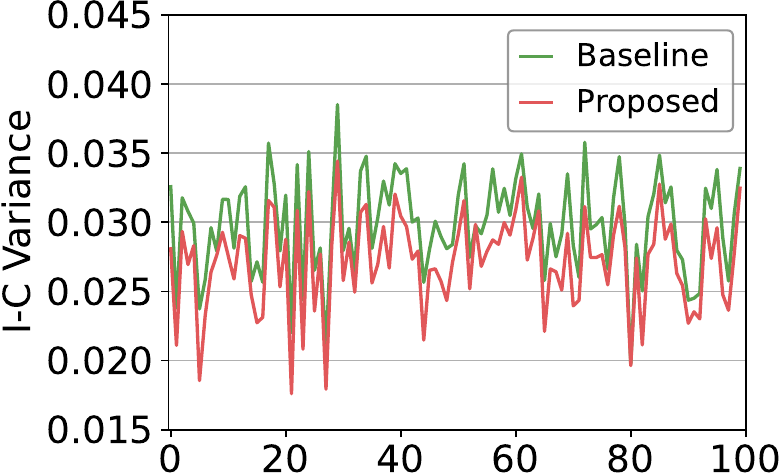}
		\caption{}
		\label{aicc}
  	\end{subfigure}
        \centering
	\begin{subfigure}{0.245\linewidth}
		\centering
		\includegraphics[width=1.0\linewidth]{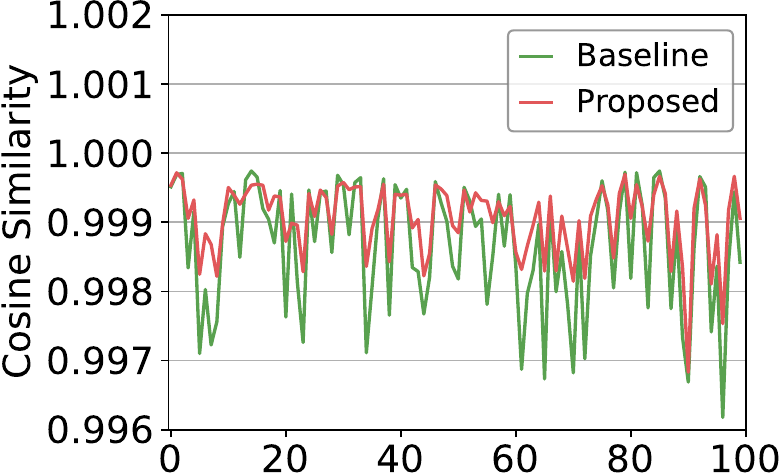}
		\caption{}
		\label{acsw}
	\end{subfigure}
 	\centering
	\begin{subfigure}{0.245\linewidth}
		\centering
		\includegraphics[width=1.0\linewidth]{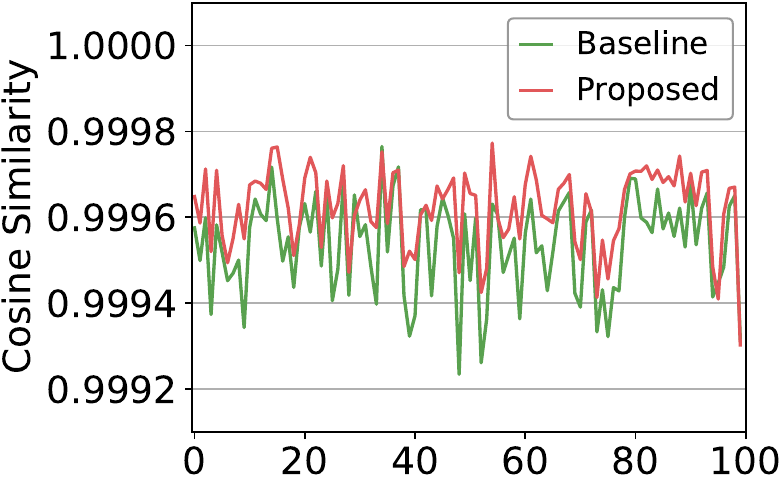}
		\caption{}
		\label{acsc}
	\end{subfigure}  
	\caption{Comparison of the difference in feature extraction of models pruned by our method and the baseline. Our method achieves more uniform feature representations. (a) Intra-cluster variance of VGG-16 through unstructured pruning. (b) Intra-cluster variance of ResNet-19 through structured pruning. (c) Cosine similarity of the means of features for each class between the training and test dataset on VGG-16 through unstructured pruning. (d) Cosine similarity of the means of features for each class between the training and test dataset on ResNet-19 through structured pruning.}
	\label{fig:a1}
\end{figure*}

\begin{figure}[htb]
        \centering
	\begin{subfigure}{0.49\linewidth}
		\centering
		\includegraphics[width=1.0\linewidth]{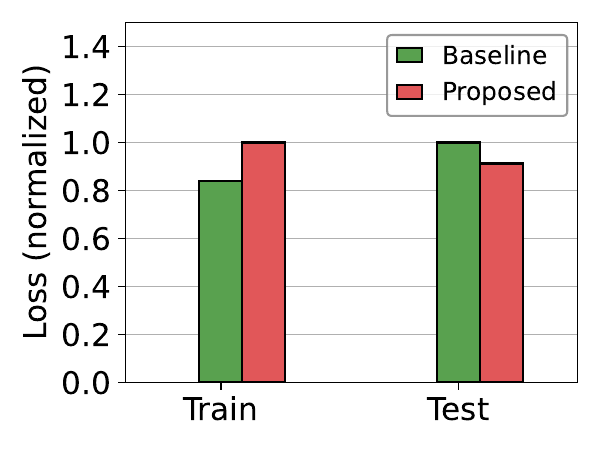}
		\caption{}
		\label{alossw}
	\end{subfigure}
 	\centering
	\begin{subfigure}{0.49\linewidth}
		\centering
		\includegraphics[width=1.0\linewidth]{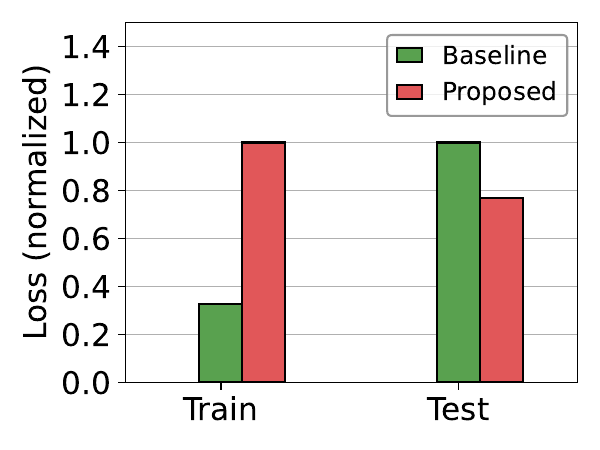}
		\caption{}
		\label{alossc}
  	\end{subfigure}
	\caption{(a) Train and test loss (normalized) on VGG16 for CIFAR100 after unstructured pruning using the proposed method and the baseline. (b) Train and test loss (normalized) on ResNet19 for CIFAR100 after structured pruning using the proposed method and the baseline.}
	\label{fig:aloss}
\end{figure}

\subsection{Feature Extraction} \label{sec:fe}
We compare the difference in feature map of models pruned by our method and the baseline. Fig. \ref{ic}, \ref{aicw} and \ref{aicc} show the intra-cluster (I-C) variance~\citep{kiang2001extending} of the features in CIFAR100 for unstructured and structured pruning. The intra-cluster variance measures the compactness of intra-cluster sample representations~\citep{oti2020new}. And it computes the sum of squared 
distance between cluster samples and the corresponding centroid: 
\begin{equation}
    \text{I-C Variance}(C_k) = \sum_{\mathbf{x}_i \in C_k}{\| \mathbf{x}_i -\mathbf{\mu}_k\|^2}, \label{ap:e1}
\end{equation}
where $\mathbf{\mu}_k$ is the mean vector. 

In our experiment, we divide intra-cluster variance by the number of samples in the cluster for the comparison. We extract feature maps before the fully connected classifier and normalize them to eliminate the influence of absolute values. According to Fig. \ref{ic}, our model learns more consistent feature representations and achieves a lower mean of variances, indicating enhanced compactness in class features. Fig. \ref{aicw} and \ref{aicc} further indicate that our model achieved lower variances in 93\% of classes on average. This improvement is consistent for both unstructured and structured pruning.

Furthermore, Fig. \ref{cs} illustrates the mean of cosine similarity of the class feature map between the training and test datasets. Fig. \ref{acsw} and \ref{acsc} further illustrate the cosine similarity of sample features in  over 98\% of classes. Our model exhibits higher similarity across almost all classes, leading to a more uniform representation of features between the train and test samples and reducing noise. Fig. \ref{accw} and \ref{accc} show the accuracies of pruned models for the training and test set after unstructured and structured pruning. Our model exhibits higher test accuracy after pruning but lower train accuracy than the baseline. We also report the training and test loss of pruned models by our method and the baseline in Fig. \ref{fig:aloss}, and the results demonstrate a similar pattern to Fig. \ref{accw} and \ref{accc}.

Based on the above observations, the baseline models suffer from overfitting during fine-tuning. The pruned models overfit the training set features, resulting in a drop in inference performance. Our method mitigates this issue by enabling the critical model to learn more consistent feature representations across the training and test sets. As a result, it reduces the noise between training and test samples, leading to improved inference performance.

\section{Conclusion}

In this paper, inspired by the critical brain hypothesis in neuroscience and the high biological plausibility of SNNs, we analyze criticality in SNNs from the perspective of maximizing feature information entropy and propose a low-cost metric for assessing neuron criticality. Incorporating this criticality into the pruning process facilitates the development of a highly efficient pruning-regeneration method. Experimental results demonstrate that our method outperforms previous SNN pruning methods in both performance and efficiency for unstructured and structured pruning. Moreover, further experiments confirm our insight that the high criticality of the pruned model improves feature extraction. An interesting future direction is to further explore the criticality in different scales of SNNs and exploit it to guide model design and high-efficiency learning.

\section*{Acknowledgements}
This work was supported by the National Natural Science Foundation of China (No. 41930110).

\bibliographystyle{elsarticle-harv}
\bibliography{ref}

\end{document}